# Design and Application of Multimodal Large Language Model Based System for End to End Automation of Road Crash Database Generation

**MD Thamed Bin Zaman Chowdhury***
Department of Civil Engineering
Bangladesh University of Engineering and Technology
Dhaka-1000
Email: zamanthamed@gmail.com

**Moazzem Hossain**
Department of Civil Engineering
Bangladesh University of Engineering and Technology
Dhaka-1000
Email: moazzem@ce.buet.ac.bd

*Corresponding Author

**ABSTRACT**
Road traffic crashes are a critical public safety issue in low-resource countries like Bangladesh, where crash data collection relies heavily on manual, incomplete police First Information Reports (FIRs). This study presents a fully automated, end-to-end crash data framework powered by Large Language Models (LLMs), moving beyond prior systems that automate only portions of the pipeline. The framework introduces two key LLM-driven autonomous agents: one that generates Python-based web scraping scripts by analyzing website structures (pagination, dynamic content, infinite scrolling) with 91.3% accuracy, and another that performs semantic-based duplicate detection, significantly outperforming traditional rule-based methods. The system consists of four integrated modules: (1) LLM-based web scraper code generation, (2) automated crash news article collection, (3) crash classification and structured data extraction, and (4) LLM-powered deduplication. A multimodal generative LLM classifies crash-related news with a 94.6% F1-score and extracts key details such as time, location, fatalities, injuries, and vehicle types. Both open-source and proprietary LLMs were evaluated to ensure adaptability in resource-limited environments. Deployed across 14 major Bangladeshi news portals over 111 days, the system processed more than 12,000 articles and identified 705 unique crash events within hours. The curated dataset revealed spatial and temporal patterns, highlighting Chittagong as the most crash-prone district and rush hours as high-risk periods. This is the first fully autonomous crash data collection system tailored to low-resource settings, enabling scalable, real-time, and policy-relevant road safety monitoring.

## INTRODUCTION

Road traffic accidents are a major public health and socio-economic challenge globally, especially in lower-income countries like Bangladesh. Worldwide, over 1.3 million people die in road crashes annually, with 93% of these fatalities occurring in low- and middle-income countries *(1)*. As a lower-middle-income country *(2)*, Bangladesh faces a severe burden. Each year, approximately 4,000 people die and around 200,000 are injured in road accidents *(3)*.

The economic toll is also substantial. Traffic crashes cost Bangladesh an estimated 5.1% of its GDP *(3)*, including immediate costs like emergency response and long-term impacts such as lost productivity and socio-economic disruption. Systematic accident analysis is essential for identifying risk factors, assessing the effectiveness of safety measures, and guiding interventions such as better traffic management, improved road design, and stricter law enforcement.

A data-driven approach allows policymakers to monitor road safety efforts and make timely adjustments. However, poor data quality remains a key obstacle in Bangladesh *(3)*, primarily due to the reliance on manual data collection and analysis.

Recent advancements in artificial intelligence, particularly Large Language Models (LLMs), offer a powerful alternative. LLMs can interpret and generate human-like text, making them suitable for automating data collection, classification, and analysis. They reduce human error, save time, and increase operational efficiency.

This study investigates the use of multimodal LLMs to automate road accident dataset generation in Bangladesh. It tackles three main challenges: the lack of reliable real-time data, the need for localized datasets tailored to Bangladesh's context, and the challenge of deploying AI solutions in resource-limited environments. The goal is to provide policymakers, planners, and public health experts with robust tools to enhance Bangladesh's road safety initiatives.

## LITERATURE REVIEW

The use of Large Language Models (LLMs) to extract structured data from unstructured accident texts has gained traction in recent years. Two broad categories of LLMs are commonly employed: bi-directional encoder-based models (e.g., BERT, RoBERTa) and uni-directional decoder-based models (e.g., GPT, LLaMA). Among these, BERT-based models have been the preferred choice for accident data extraction, though generative models are emerging as viable alternatives.

Study in *(4)* leveraged Named Entity Recognition (NER) with transformer models like BERT, RoBERTa, and XLNet to extract entities such as accident locations, victim details, causes, and fatalities from Indian newspaper articles. Fine-tuning on a custom dataset enabled high accuracy, though narrative variation in articles posed challenges.

Authors in *(5)* developed an automated platform, ARTCDP, to extract structured data from Chinese road accident reports. Using BERT-based classification, the system filtered relevant reports and extracted 27 variables, including date, time, and location. While most variables saw over 80% accuracy, performance on key fields like date and non-fatal injuries dropped below 70%.

Researchers in *(6)* fine-tuned BERT on labeled police accident narratives. Their BERT-based classifier significantly outperformed traditional approaches (e.g., TF-IDF with SVM), effectively identifying subtle linguistic cues tied to accident severity, such as high speed or driver negligence.

Authors in *(7)* combined BERT embeddings with a Recurrent Convolutional Neural Network (RCNN) to capture both global and local text features. Their model addressed class imbalance—common in accident data—by using focal loss and resampling. This hybrid architecture outperformed standalone BERT or CNN models, particularly in detecting rare accident types.

The study in *(8)* proposed a two-step NLP pipeline to process accident-related tweets. First, a classifier filtered relevant tweets. Then, a BERT-based joint model performed slot-filling to extract details such as location and event descriptions. Their multitask model, trained on datasets from Belgium and the Brussels region, achieved over 95% F1-score on both classification and entity extraction tasks.

In study *(9)*, authors collected 26,226 accident-related tweets from Australia and used GPT-3.5 to annotate each with 15 features through prompt engineering. These included both classification (e.g., is there a fatality?) and extraction tasks (e.g., involved vehicles). A smaller model, GPT-2, was then fine-tuned under a multitask learning (MTL) setup, combining classification and generative tasks. This fine-tuned GPT-2 outperformed baselines, including zero-shot GPT-4 mini and XGBoost, achieving an average classification accuracy of 85%.

In study *(10)*, authors explored prompt-based performance of GPT-3.5 Turbo and Meta's LLaMA-3 (8B and 70B). Using different inference strategies such as zero-shot (ZS), few-shot (FS), and Chain-of-Thought (CoT) prompting, they observed significant accuracy improvements without fine-tuning. For instance, GPT-3.5 with ZS+CoT achieved 49.3% macro-accuracy compared to ~35% with naive ZS. In some CoT-enhanced settings, the model achieved up to 100% accuracy on certain severity classes. Their results highlight how prompt design alone can drastically boost performance.

Despite the advancements, several gaps remain. Most scraping scripts are tailored to specific websites, demanding manual effort and technical expertise. A more automated solution that generates scraping code based on inputs like newspaper name and URL would make data collection scalable.

Additionally, consolidating news from multiple sources while avoiding duplication is a labor-intensive task when handled manually. A system capable of automatically identifying and removing duplicates can significantly reduce human workload.

Notably, none of the reviewed studies focus on Bangladesh—a country with unique challenges such as scarce resources, inconsistent accident reporting styles, and a lack of robust NLP infrastructure. While countries like China, Australia, and Albania have seen automated accident data pipelines, Bangladesh has yet to develop similar systems. Given the infrastructural limitations, approaches focusing on prompt engineering instead of fine-tuning offer a practical pathway for implementation in such low-resource settings.

The specificity of Bangladeshi news formats, local languages, and traffic reporting norms means that existing models cannot be directly transferred. Developing prompt-based LLM solutions tailored to local contexts—without the need for resource-intensive fine-tuning—could be a breakthrough in accident data automation for underdeveloped countries.

## METHODOLOGY

### *Flowchart of the study*

**Figure 1** presents the methodological framework of this study. The study began with a survey of Bangladeshi online newspapers. Each site's robots.txt file was checked for scraping compliance, followed by structural analysis to categorize sites by pagination style (e.g., traditional, infinite scroll, dynamic loading). A set of primary news websites was selected, with 80% of the data used for system calibration and 20% for validation.

The first component, Automated Scraping Code Generation Agent, employed prompt-engineered LLMs to generate scraping scripts. Techniques like few-shot prompting and in-context learning *(11)*, along with HTML batching and token management, enabled scalable, adaptive scraping across diverse formats. A CodeAct agent *(12)* handled execution and debugging of generated code.

Next, the Automated Duplicates Removal Agent used LLM-based semantic similarity to eliminate redundant reports, ensuring data integrity. Then, the Classification and Information Extraction Agent categorized articles into accident and non-accident types and extracted structured details from the former. The system was calibrated using the design set and evaluated on a separate validation set, including expert-verified labels for benchmarking. Finally, the structured data powered an Application Layer that visualized accident trends in Bangladesh through heat maps and temporal charts, aiding in policy and safety analysis.

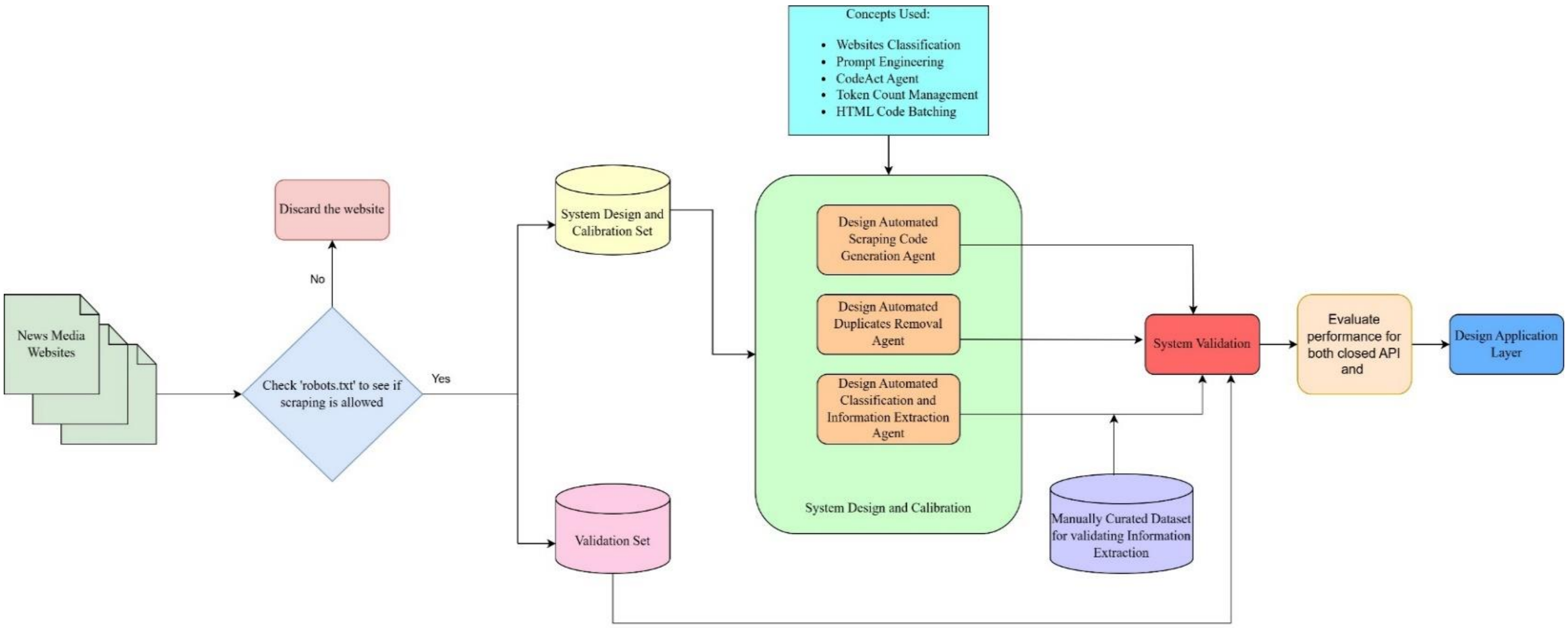

**Figure 1 Study Flow Chart**

*Legal Compliance of Web Scraping*
We strictly followed the ACM Code of Ethics and Professional Conduct *(13)*. Our system includes a mechanism to check a website's robots.txt before scraping; if scraping is disallowed, the process is aborted. We collected only publicly available news articles for academic research, respecting site restrictions and avoiding authentication-protected pages. To prevent server overload, we added delays between requests. Only essential data (title, publish date, body text) was extracted—no user data or media assets. The entire process prioritized ethical and legal compliance, aligning with ACM guidelines on harm avoidance (1.2) and privacy respect (1.6).

*LLMs used and testing environment*
A total of four LLMs are tested in this study. To ensure reproducibility, we compared performance of some open-source LLMs besides a closed API based one. As a result, our system can be used both using an API or locally in a capable enough machine. **Table 1** summarizes the LLMs used in this study.

**TABLE 1 LLMs Used in the Study**

| LLMs Tested | Type of LLM | Open-source | Parameter Count | Temperature | Top P | Top K |
|---|---|---|---|---|---|---|
| Gemini-2.5-flash | Multimodal | No | -- | 1 | 0.95 | 64 |
| Gemma-3-27B-it-unsloth-bnb-4bit | Text | Yes | 16.1B | 1 | 0.95 | 64 |
| Qwen-3-32B-AWQ | Text | Yes | 5.73B | 1 | 1 | 50 |
| Gemma-3-12B-it | Multimodal | Yes | 12.2B | 1 | 1 | 50 |

We used LLMs with varying parameter sizes to determine the minimum computational resources required. While Google has not disclosed the parameter count of Gemini-2.5-flash *(14)*, its support for up to 1 million tokens suggests it exceeds models like Gemma-3-27B-it-unsloth-bnb-4bit *(15)*, which supports 128k tokens. Gemini-2.5-flash, a closed API model, was used across all pipelines. Open-source models like Gemma-3-27B, Qwen-3-32B-AWQ *(16)*, and vision-capable Gemma-3-12B-it supported tasks including code generation, classification, and deduplication. During the calibration phase, we used Gemini-2.5-flash with few-shot and in-context learning, embedding examples from the calibration set in

prompts—potentially boosting calibration accuracy over validation. All models were ultimately evaluated using the separate validation set.
All the testing for the open-source LLMs were done locally in a single NVIDIA L40S GPU with 48GB VRAM. The system had 128GB of RAM. It had 16 vCPUs each running on average 2.65GHz clock speed. The tests were done in the following environment: Transformers: 4.52.4 *(17)*, Platform: Linux, Torch: 2.7.0+cu126 *(18)*. CUDA: 8.9. CUDA Toolkit: 12.6 *(19)*. Triton: 3.3.0 *(20)*.

*Overall System Architecture*

There are several components working together in the system. The system utilizes multiple LLM agents to achieve its objectives. These components work under four subsystems. These are: Web Scraping Code Generation, News Collection, News Classification and Feature Extraction from Accident Reports and lastly, Duplicate News Detection and Data Consolidation.
**Figure 2** illustrates the overall architecture of the system and how the subsystems interact with each other to achieve the objective.

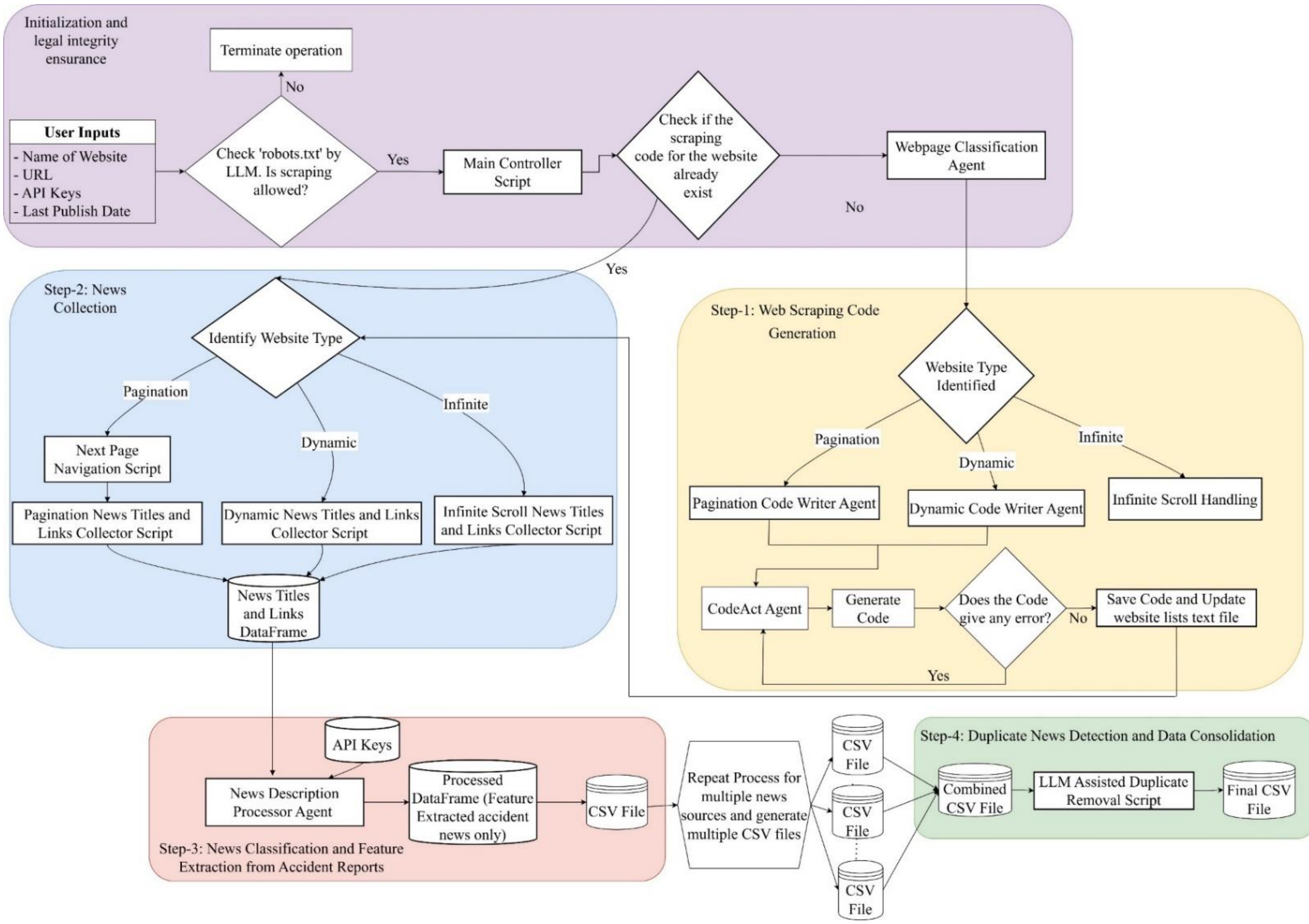

**Figure 2 System Architecture**

*Automated Webpage Classification*

The authors first studies all the available online newspapers and classified them. Then the authors chose three types of websites to design the scraping code generator as those three types cover 92% of the country's online news sites. The result of the study is included in the results section.

The Webpage Classifier Agent is designed to analyze a given webpage and determine its loading mechanism. Specifically, it classifies the webpage as one of the following three types:

i. Infinite Scroll – The webpage continuously loads more content as the user scrolls down.

ii. Dynamic – The webpage loads more content when a "Load More" button (or equivalent) is clicked.

iii. Pagination – The webpage uses numbered page navigation (1, 2, 3, etc.) or "Next Page" buttons for content browsing.

These types were identified by extensively studying the website structures of the calibration set. Most of the websites fall under one of these three categories. To perform the classification, the agent captures screenshots of the website automatically using Selenium *(21)*. Then it sends the screenshot along a detailed prompt to the LLM. **Table 2** is an example of the prompts.

**TABLE 2 Visual Input and Prompt Template of Website Classification Agent**

| **Website Classification** |
| --- |

| | |
|---|---|
| Prompt | 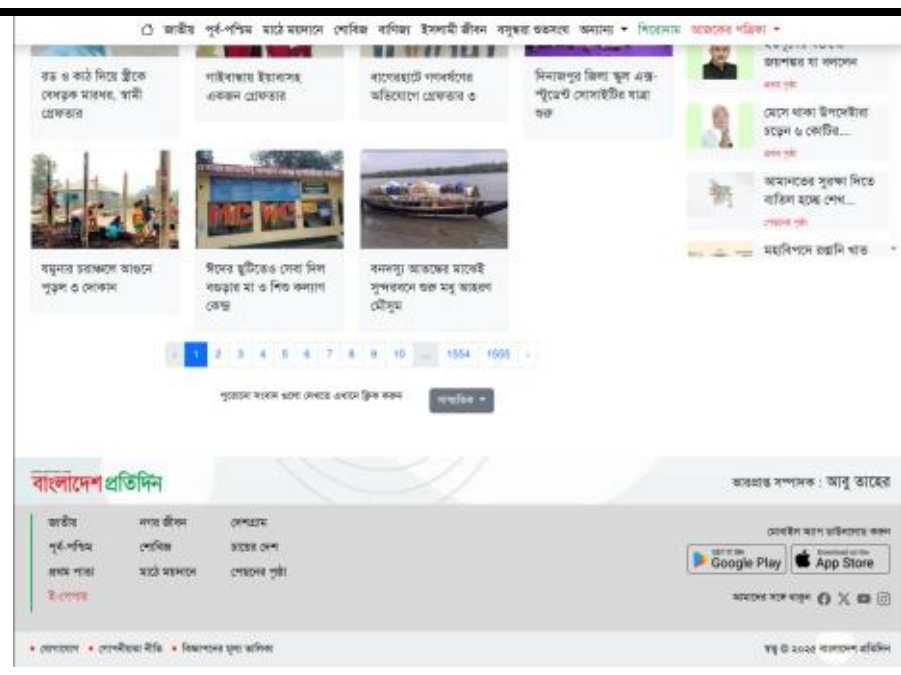<br>In this webpage screenshot, can you see any numbered next page buttons (1,2,3 etc.), next page navigation button or 'load more'/আরও/আরো type of button?<br>The buttons may be in bangla as the screenshot can be of a bengali or english newspaper website.<br>Your answer should be in this format: <answer> No <answer> or <answer> Yes <answer> |
| Response | <answer> Yes </answer> |
| Prompt | 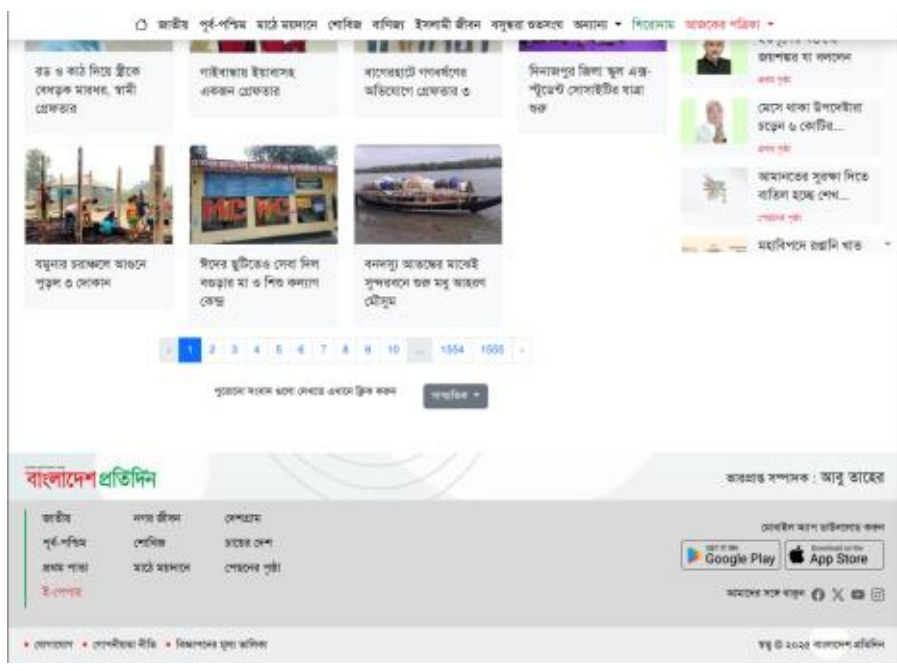<br>Classify the website based on the screenshot or url or html code as either 'dynamic' or 'pagination'.<br>A 'dynamic' website uses buttons like 'Load More' to load additional content dynamically without changing the page.<br>A 'pagination' website has numbered pages (e.g., 1, 2, 3) at the bottom or in the urls to navigate between different pages.<br>Your response should be in the format: <type>dynamic<type> or <type>pagination<type><br>-----------------------------------<br>URL of the website: {url}<br>-----------------------------------<br>HTML code of the website:<br>{HTML_code} |
| Response | <type>Pagination</type> |

*Automated Scraping Code Generator*

The system assigns scraping tasks based on website type to specialized agents: the Pagination Code Writer Agent handles sites with page numbers or navigation buttons, the Dynamic Code Writer Agent manages sites with dynamically loaded content, and Infinite Scroll Handler deals with endless scrolling pages without generating individual code.

At the core of the Pagination and Dynamic Agents is the CodeAct Agent, which extracts Python code generated by an LLM—identified within triple backticks—and saves it to a file for execution. If the code fails, the agent initiates a debugging loop by prompting the LLM with detailed error feedback, iteratively refining the script until successful execution is achieved.

To manage large HTML documents, the system uses Selenium and BeautifulSoup *(22)* to download full page content, then divides it into chunks to stay within LLM token limits. Token count is precisely tracked using Google Vertex AI's tokenizer *(23)*. Each HTML chunk is analyzed individually. If it contains critical elements like "load more" buttons or pagination links, processing halts. If not, the system continues with the next chunk, ensuring comprehensive parsing without exceeding token constraints. This approach ensures adaptive, error-resilient scraping across diverse site architectures.

The generated code will be used to collect news automatically from the respective news media website. Overall, this workflow combines automated code generation, error handling through iterative debugging, and efficient resource management techniques to produce customized web scraping scripts tailored to diverse website architectures.

*Automated Classification and Information Extraction*

The system processes a DataFrame (merged_df) containing URLs to news articles. For each URL, webpage content is retrieved using the Requests library with a custom User-Agent header to mimic a browser and increase successful access *(24)*. BeautifulSoup then parses the HTML, extracting clean text for analysis.

Next, the system leverages a large language model (LLM) through the report_processing_agent function. Using carefully designed prompts, the LLM classifies articles as either General (non-road incidents or summaries) or Specific (detailed descriptions of a single road traffic accident). For Specific articles, the LLM extracts structured data points: publish date, accident date and time, number of deaths and injuries, accident location, road type, pedestrian involvement, vehicle types, and district. These outputs follow a fixed format, separated by <sep> tags for easy parsing.

The script features robust error handling and API key rotation. If a network or API-related issue occurs, it retries with alternate keys and includes wait times to mitigate rate-limiting or bans. This ensures reliability when processing large article volumes.

Finally, the LLM outputs are reintegrated into the original DataFrame. Articles labeled "General" are filtered out, and "Specific" entries are split into structured columns using the <sep> delimiter. Missing or inconsistent fields are clearly flagged for review. The result is a clean, structured DataFrame containing detailed, article-level accident data—ready for downstream tasks such as analysis, visualization, or reporting.

*Automated Duplicate Removal and Data Consolidation*

This agent filters a news dataset to retain only detailed reports of specific road accidents. We introduce an LLM powered deduplication algorithm and compared it with traditional rule based deduplication. It can be demonstrated that LLM powered deduplication can achieve better accuracy than rule based deduplication. The results are shown in results section. The following pseudocode denotes the algorithm used in this step:

**Algorithm:** Sequential Deduplication of News Articles using LLM Responses

**Input:** Dataset (duplicates_df) containing columns: 'District', 'Accident Date', 'Publish Date', 'News Title', 'Description'.

**Output:** Filtered dataset (df_unique) containing only unique news articles.

1. Check if dataset duplicates_df is empty:
2.     If empty, terminate the algorithm.
3.     Else, proceed to step 4.
4. Group the dataset by columns 'District' and 'Accident Date'.
5. For each group created in step 4, perform steps 6 to 27:
6. Initialize two DataFrames:
7.     df_unique_group as empty (to store unique articles).
8.     candidates containing all rows from the current group.
9. While candidates is not empty, repeat steps 10 to 26:

10. Select a "base news" article:
11. If df_unique_group is empty:
12. Select the first article in candidates as "base news."
13. Move this article from candidates into df_unique_group.
14. Else:
15. Select the most recently added article in df_unique_group as the "base news."
16. Check if candidates DataFrame is empty:
17. If empty, exit loop (proceed to step 27).
18. Else, continue to step 19.
19. Use the LLM to compare the "base news" article with each candidate article and obtain boolean flags:
20. True means the candidate is a duplicate of the base news.
21. False means the candidate is unique.
22. Remove all articles from candidates identified as duplicates.
23. Check if any candidates remain:
24. If none remain, exit loop (proceed to step 27).
25. Else, continue to step 26.
26. Move the first remaining non-duplicate candidate from candidates into df_unique_group.
27. After processing all groups from step 4, combine all resulting df_unique_group DataFrames into a temporary DataFrame called df_unique_temp.
28. Repeat the deduplication logic by grouping df_unique_temp by 'District' and 'Publish Date', following the same iterative steps (6 to 26) to remove any additional duplicates.
29. Combine all resulting groups after this second pass into the final deduplicated DataFrame df_unique.
30. Output the final deduplicated dataset (df_unique).

The automated deduplication algorithm offers a resource-efficient and scalable solution for filtering and refining large-scale road accident news datasets. Additionally, this two-stage grouping serves as an effective safeguard against potential errors in ‘Accident Date’.

## RESULTS AND DISCUSSIONS

*Survey on webpages of Bangladeshi newspapers*

We conducted a survey on the available websites of Bangladeshi newspapers. Then, we studied the HTML codes of the websites carefully and classified these websites based on their pagination techniques. The results are shown in **Figure 3**.

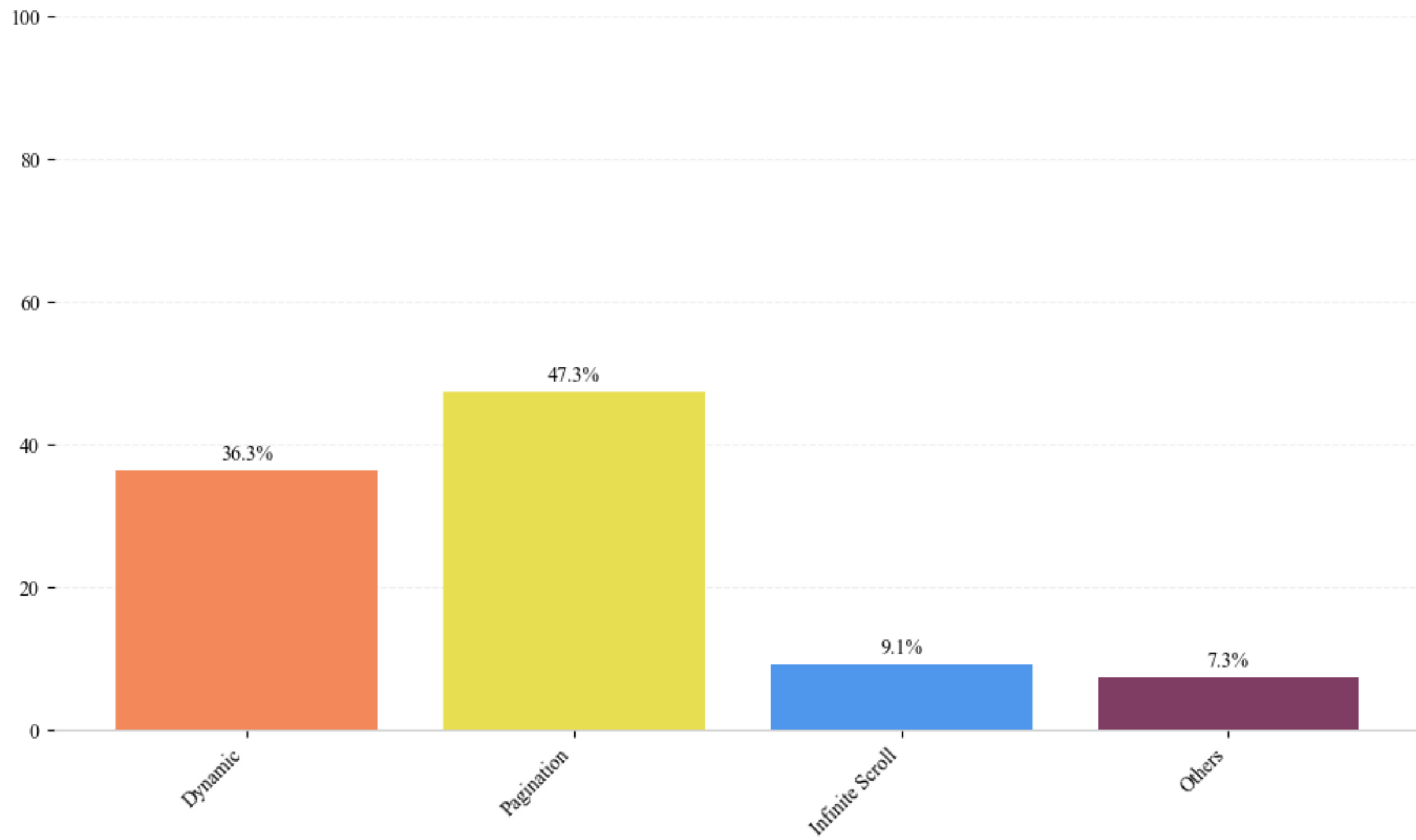

**Figure 3 Newspaper Webpage Classification of Bangladesh**

It can be clearly demonstrated that 92.7% of websites follow one of three pagination techniques: ‘Dynamic’, ‘Pagination’ or ‘Infinite Scrolling’. A very small percentage of websites use other techniques such as- Search based or hybrid methods. Since most webpages utilize one of the three methods, we designed our system to scrape news from these webpages only.

*Performance of LLMs in Website Classification Task*

Classifying websites require vision capability as the LLM classifies the webpage based on its screenshot. We compared Gemini-2.5-flash and Gemma-3-12B-it for this task. The first one is a closed API based LLM whereas the second one is an open-source LLM. Their performance is compared in **Figure 4**.

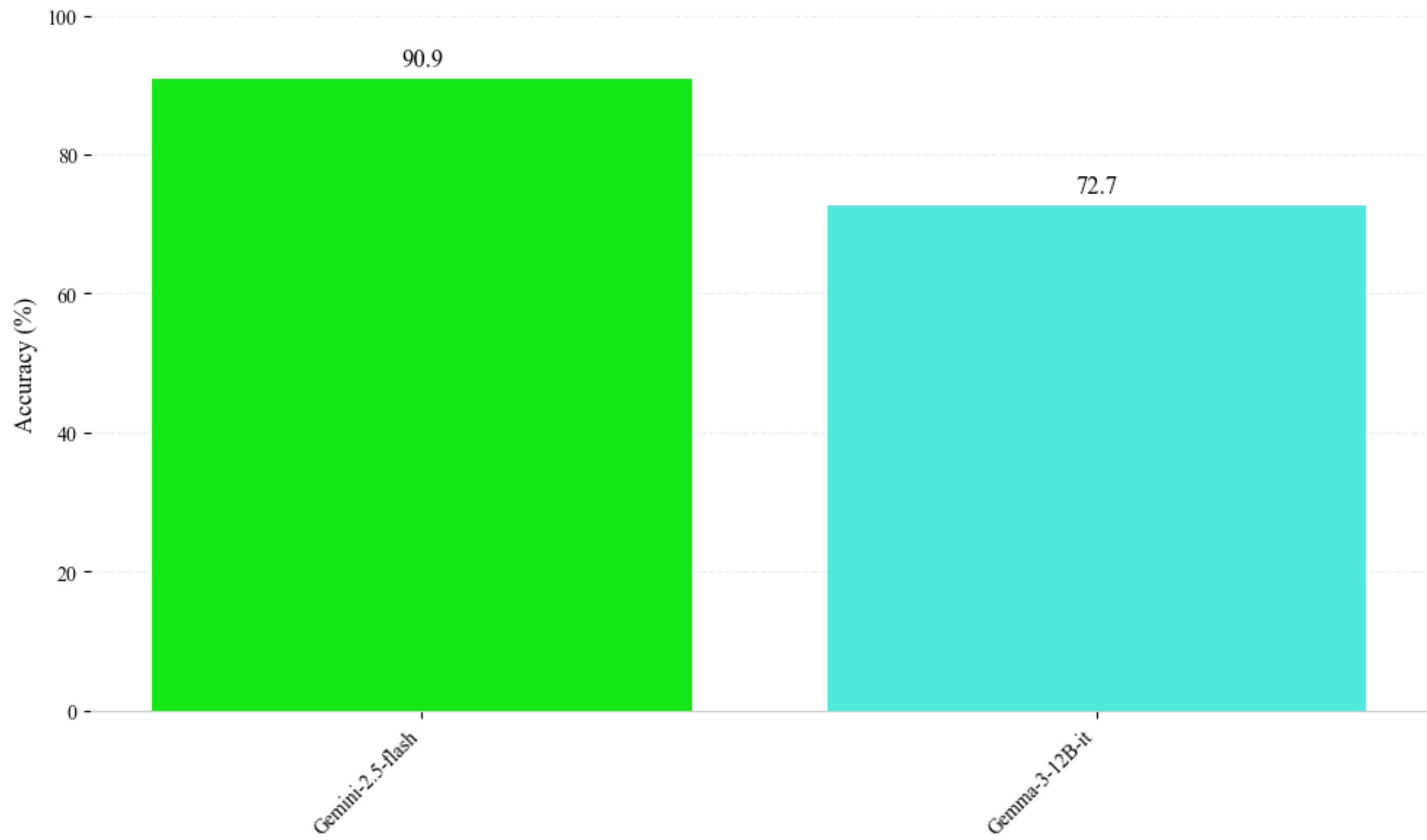

**Figure 4 Accuracy of vision LLMs in webpage classification task.**

So, it can be seen from that the model with higher parameter count benefits more in case of vision understanding. As a result, accuracy in classification of Gemini-2.5-flash is higher than Gemma-3-12B-it.

*Performance of LLMs on Automated Scraping Code Generation Task*

The Performance of web scraping component was measured by randomly choosing a total of 54 websites and using the component to collect news report automatically. Again, 80% websites were used for system development and calibration and rest of the 20% websites were used for validation.

The overall performance is shown in **Figure 5**.

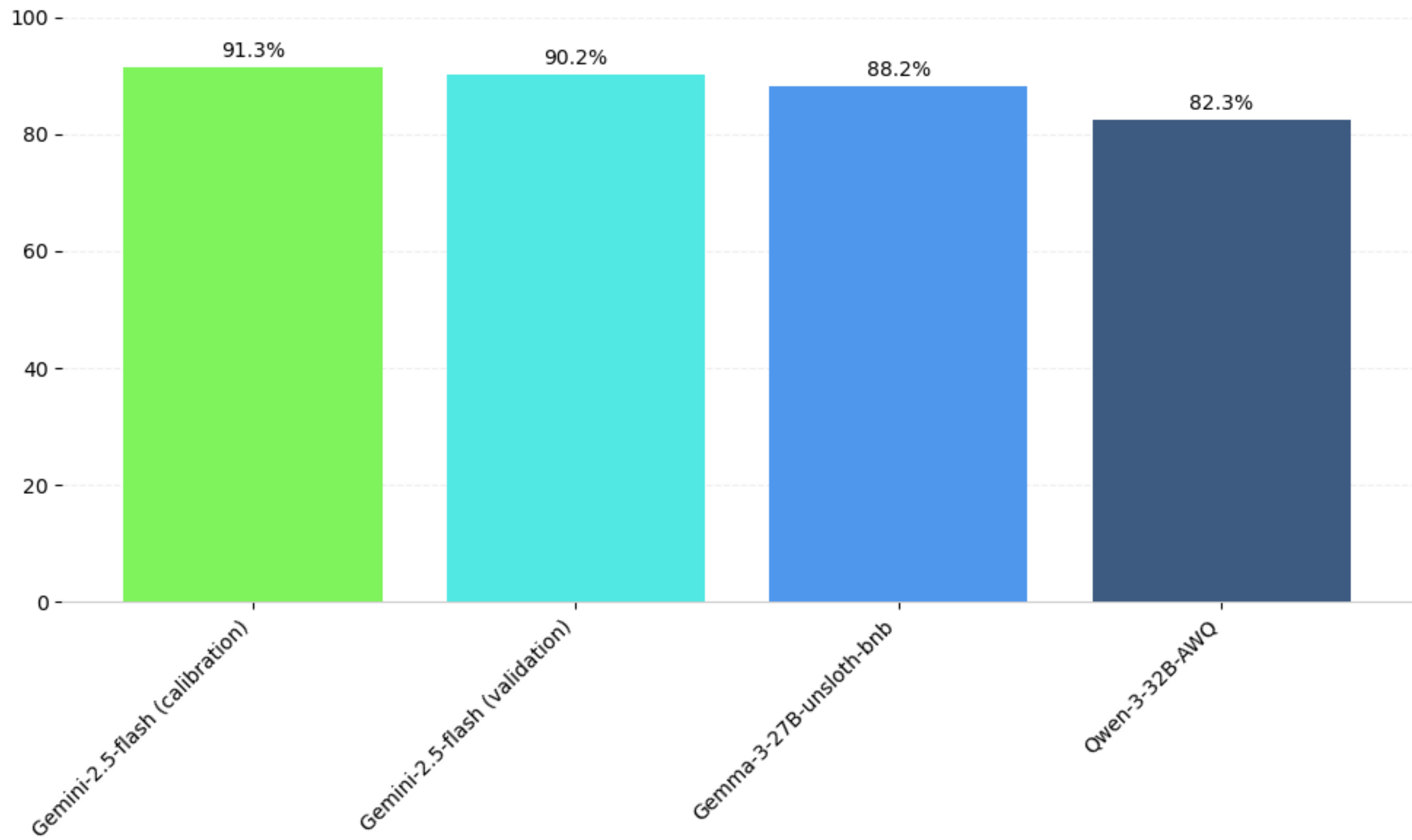

**Figure 5 Success Rates of Different LLMs on Scraping Code Generation task**

We also measured how much time it takes for each LLM to complete this task. This result is shown in **Figure 6**.

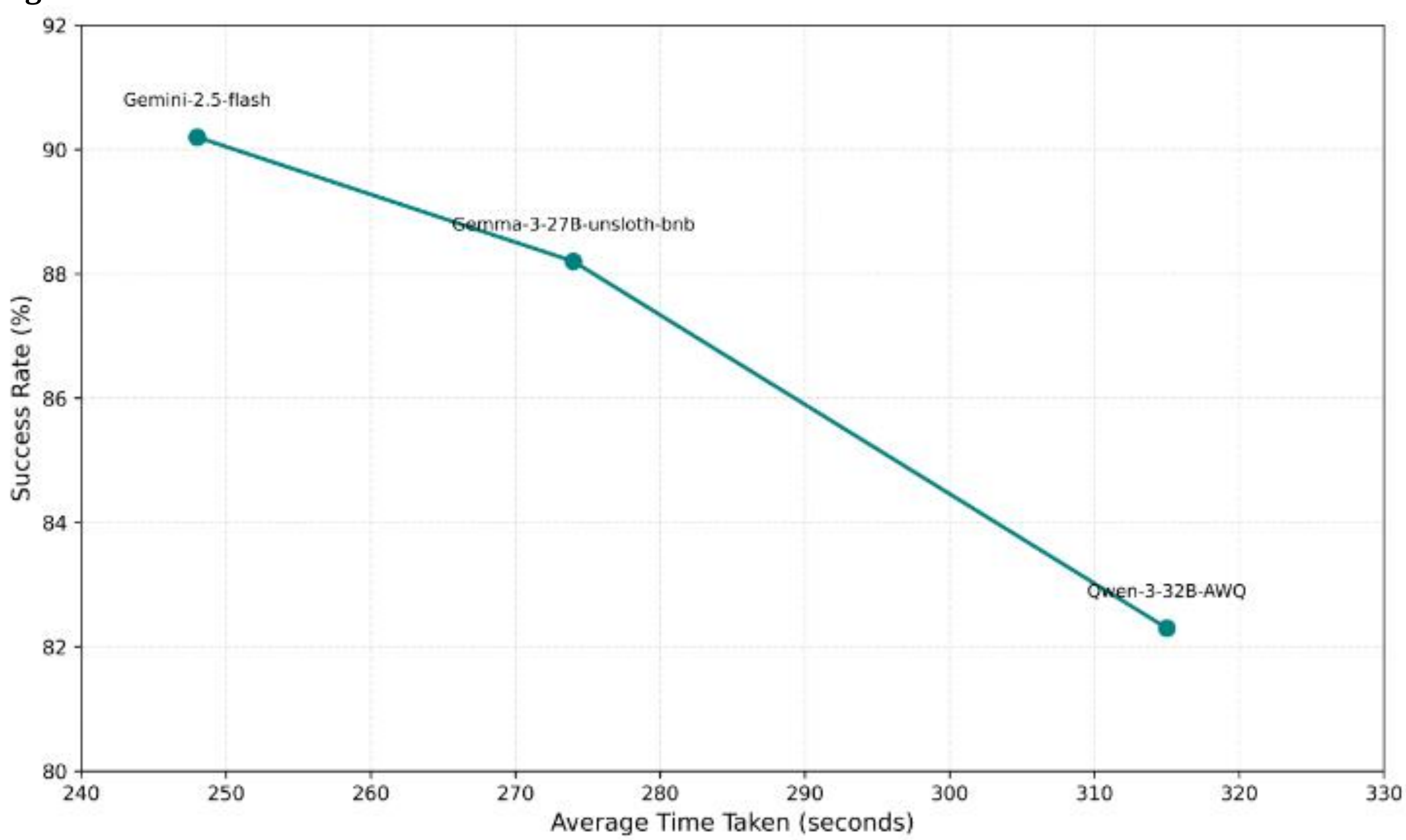

**Figure 6 LLM Success Rate on Code Generation vs. Average Time Taken**

It can be clearly seen that the higher the parameter count of the LLM, the more successful that LLM is in code generation task. Qwen-3-32B took more time than Gemma-3 despite being lower parameter count because it generated more tokens to complete this task than Gemma-3 which consumed more time.

*Accuracy of LLM on News Classification*

A manually curated dataset comprising of 268 news was used in this stage. 45% data was used for system design and calibration and 55% data was used for validation.

The confusion matrix in this task is defined in **Table 3**.

**TABLE 3 Confusion Matrix**

| | Predicted as Accident News | Predicted as Non-Accident News |
|---|---|---|
| Actually Accident News | True Positive | False Positive |
| Actually Non-Accident News | False Positive | False Negative |

The F1-score is calculated by the formula:

$$F1 = \frac{2 \times Precision \times Recall}{Precision + Recall} \quad (1)$$

Where,

$$Precision = \frac{TP}{TP+FP} \quad (2)$$

$$Recall = \frac{TP}{TP+FN}$$

(3)

Results are shown in **Table 4**.

**TABLE 4 Evaluation Metrics for Report Classification Task**

| | **Gemini-2.5-flash (Calibration)** | **Gemini-2.5-flash (Validation)** | **Gemma-3-27B-it-unsloth-bnb-4bit** | **Qwen-3-32B-AWQ** |
|---|---|---|---|---|
| Precision | 0.987 | 0.984 | 0.970 | 0.970 |
| Recall | 0.962 | 0.909 | 0.992 | 0.992 |
| F1 | 0.974 | 0.946 | 0.981 | 0.981 |

Here, it can be seen that all the models perform satisfactorily in report classification task. All F1-scores are at least 95% or above. The calibration and validation accuracies are identical indicating good generalization in report classification task.

*Accuracy of LLM on Accident Information Extraction*

The accuracy of each LLM in information extraction tasks is discussed in this section. The accuracy for each attribute in this task is defined as:

$$Accuracy = \frac{Total\ Correct\ Answers\ for\ Attribute}{Total\ Number\ of\ Data\ Points\ for\ Attribute} \times 100\% \quad (4)$$

**Figure 7** provides the system calibration accuracy which uses Gemini-2.5-flash throughout 8 attributes.

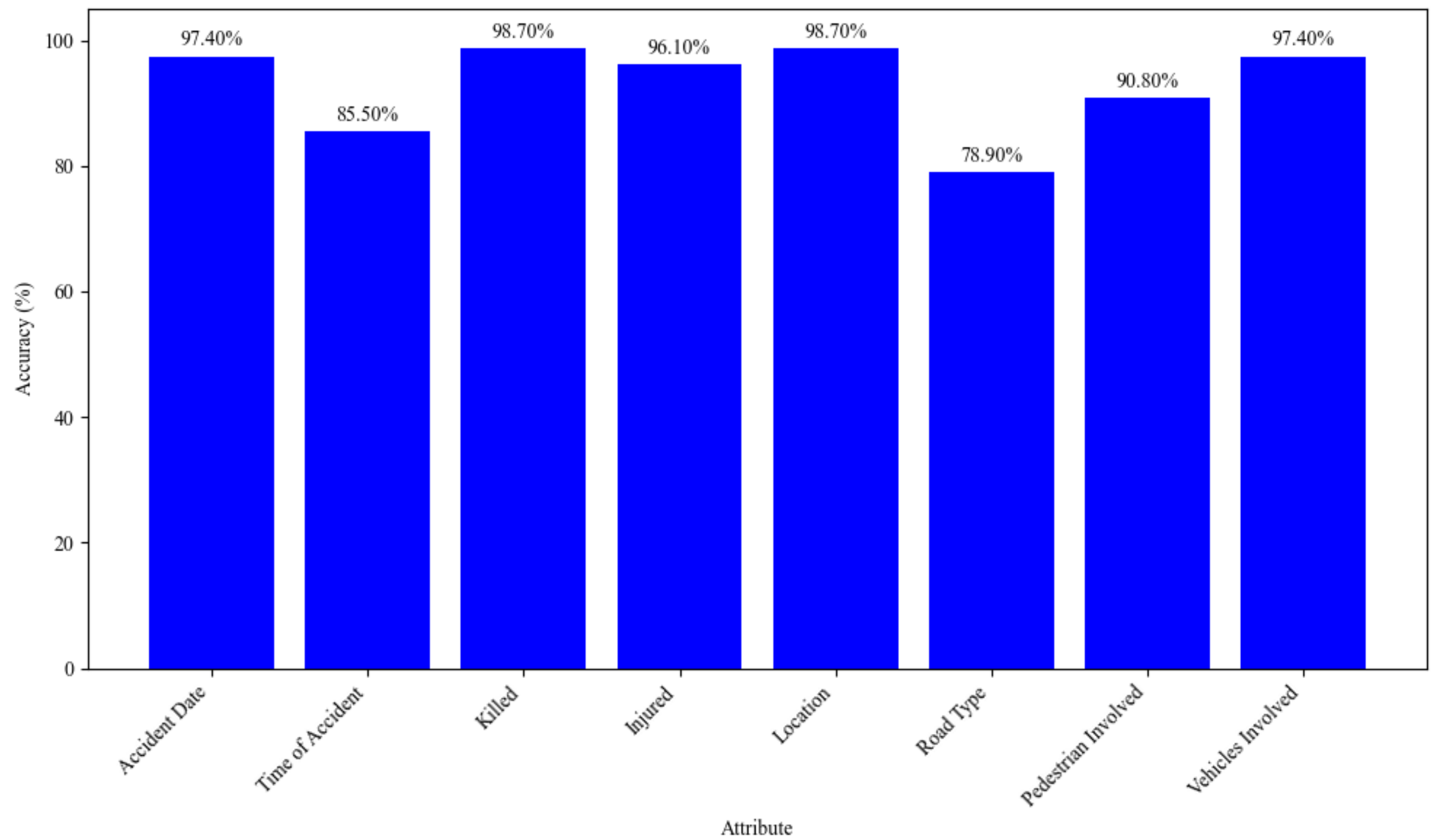

**Figure 7 Attribute wise System Calibration Accuracy in Information Extraction (Gemini-2.5-Flash)**
In the calibration phase, most attributes achieve over 90% accuracy except for 'Accident Date' and 'Road Type'. In some cases, 'Accident Date' requires reasoning as newspapers sometime mention 'Yesterday', 'Last Friday' etc., terms instead of reporting 'Accident Date' directly and LLMs sometimes struggle to do this reasoning. In case of 'Road Type', the LLM sometimes mention this attribute as 'not mentioned' if the road type is not present in the report very explicitly.

**Figure 8** provides the system validation accuracy which uses Gemini-2.5-flash throughout 8 attributes.

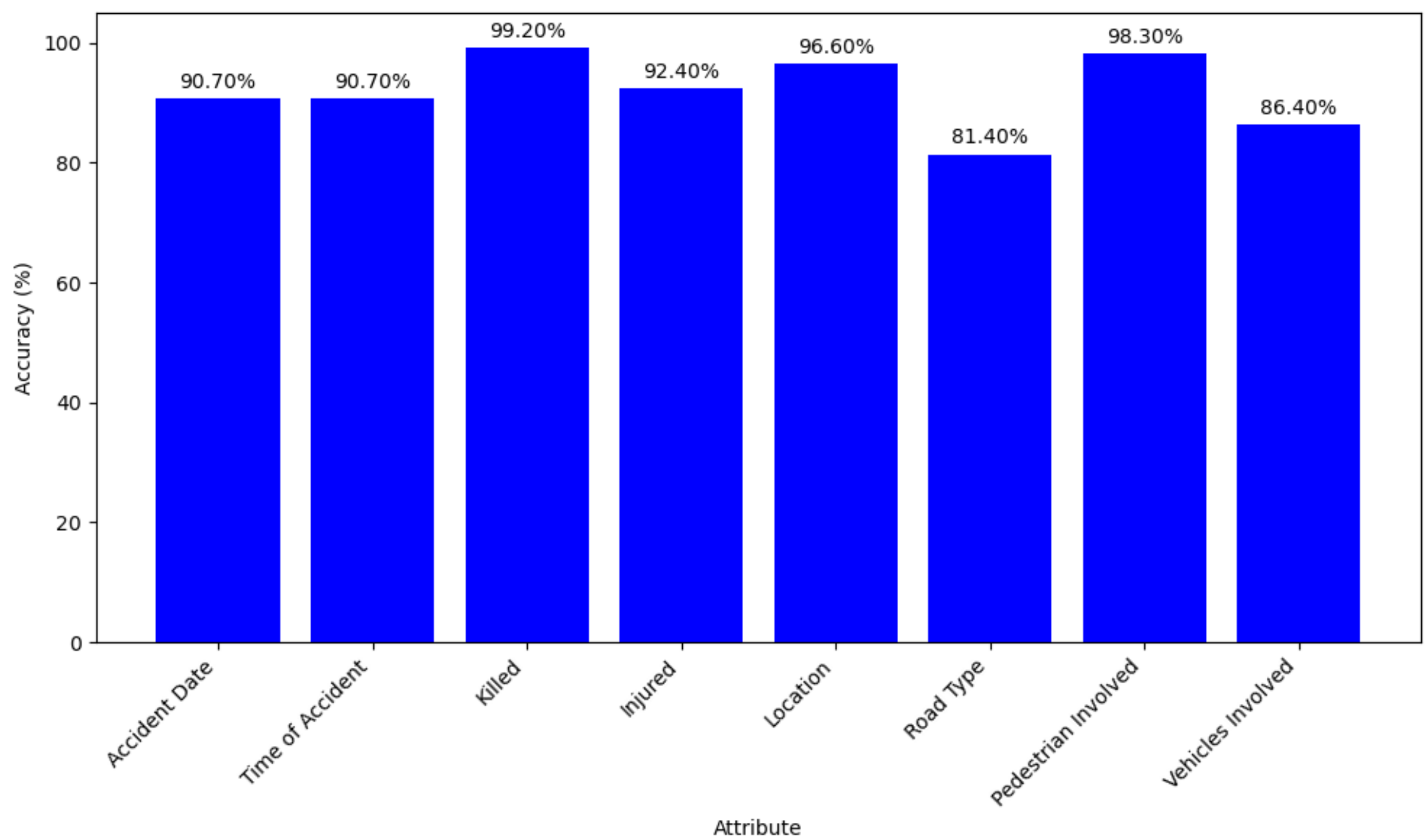

**Figure 8 Attribute wise System Validation Accuracy in Information Extraction (Gemini-2.5-Flash)**

During the validation phase, the accuracy of the attribute 'Vehicles Involved' dropped below 90%. We found that when a lot of vehicles involve in an accident or multiple accident reports come under a single report, the LLM sometimes struggle to properly capture all the vehicles involved in the accident.

**Figure 9** provides the accuracy of Gemma-3-27B-unsloth-bnb-4bit in this task.

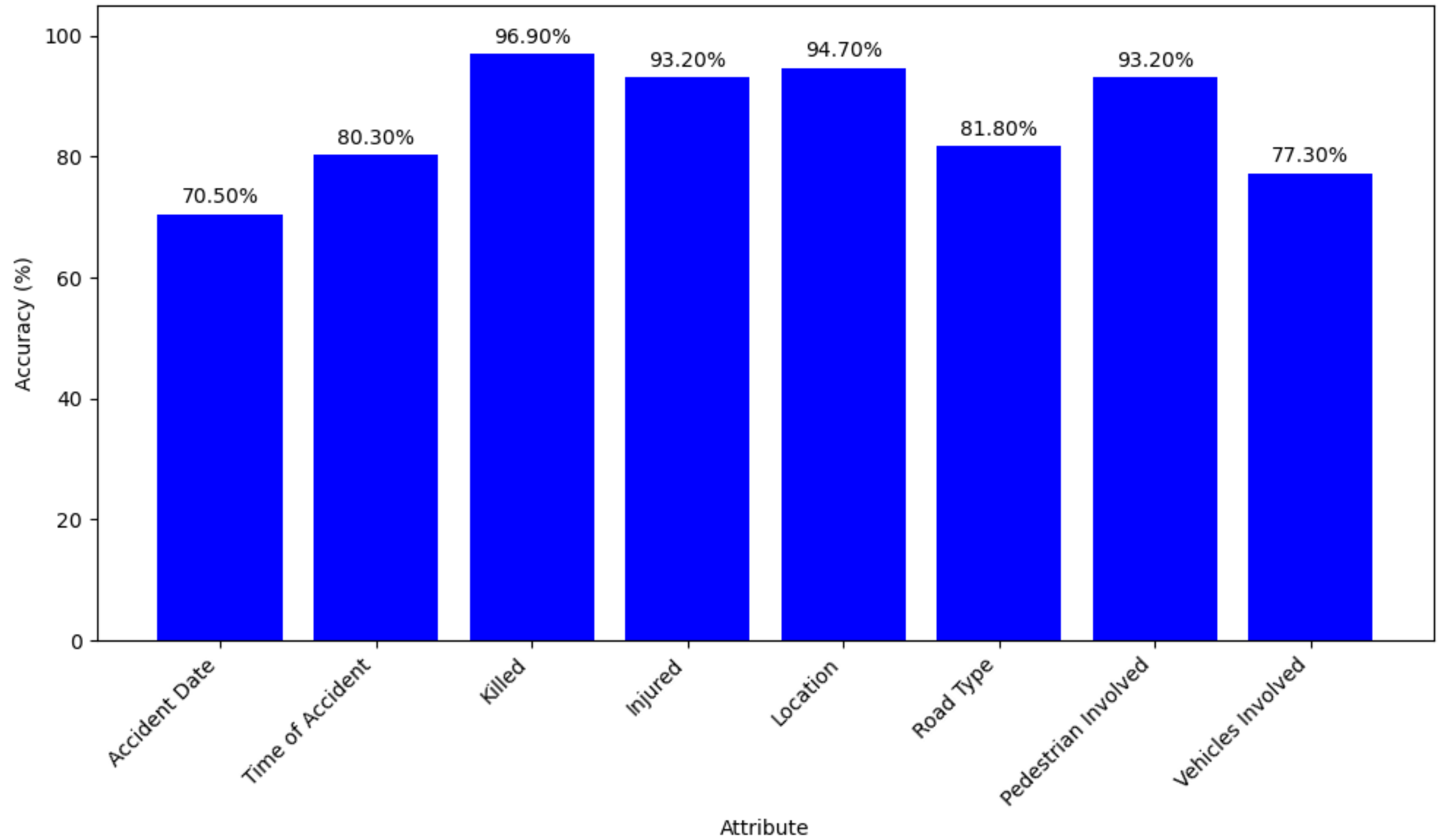

**Figure 9 Attribute wise System Validation Accuracy in Information Extraction (Gemma-3-27B-unsloth-bnb-4bit)**

Here, we can see that accuracy drops significantly in 'Accident Date', 'Time of Accident', 'Road Type' and 'Vehicles Involved'. This LLM has a lower parameter count than Gemini-2.5-flash and so, it fails to perform similarly in some attributes.

Finally, **Figure 10** provides the accuracy of Qwen-3-32B-AWQ in this task.

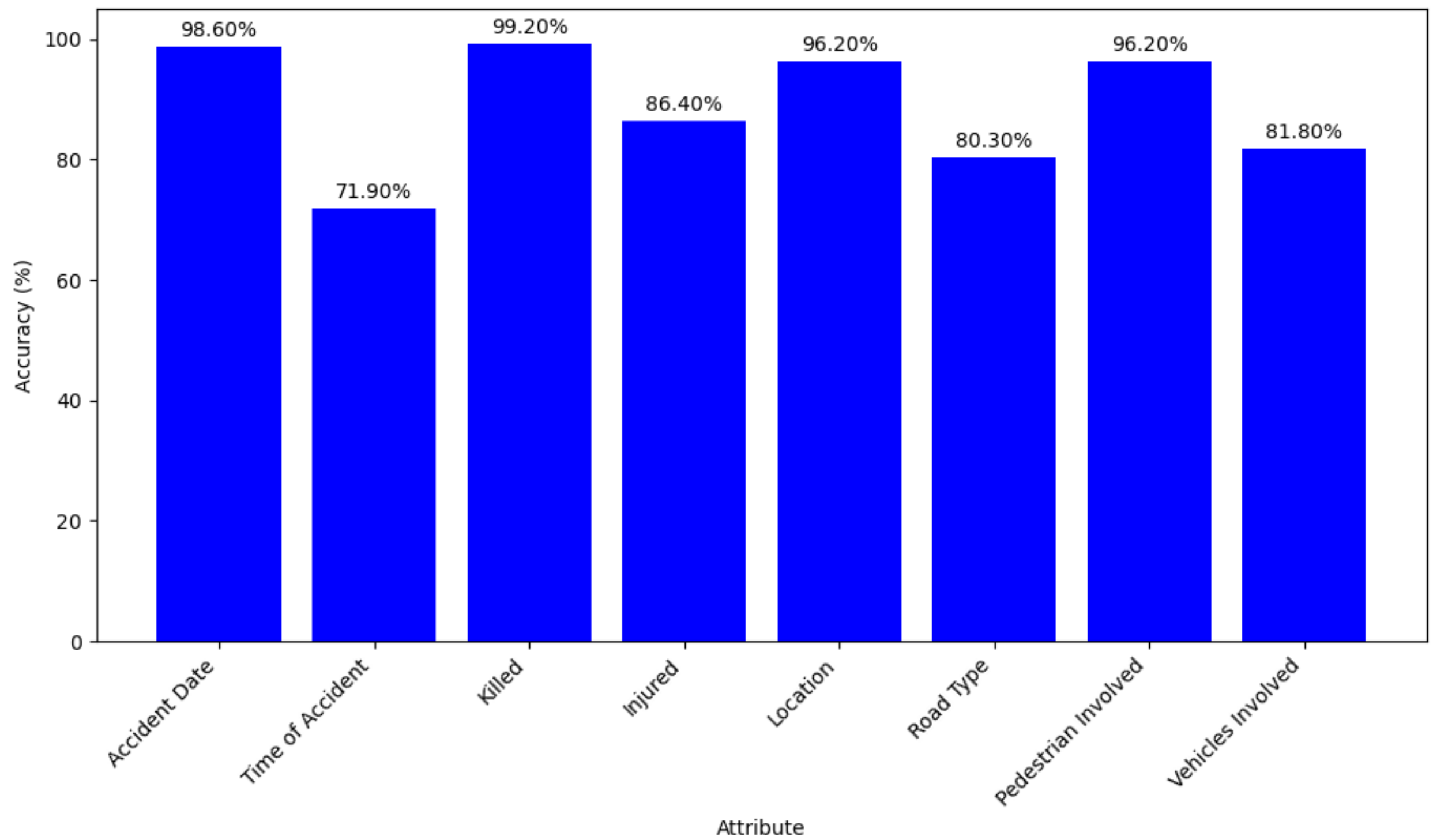

**Figure 10 Attribute wise System Validation Accuracy in Information Extraction (Qwen-3-32B-AWQ)**

Again, this LLM also does not perform satisfactorily in attributes- 'Time of Accident', ''Killed', 'Road Type' and 'Vehicles Involved'. The reason can be attributed to lower parameter count again. But a surprising matter can be seen that this LLM perform excellently in 'Accident Date' attribute where other LLMs did not perform well. The reason might be, Qwen-3-32B-AWQ is a model that is particularly tailored to handle math problems very well. Finding 'Accident Date' from news reports and 'Publish Date' is essentially a math problem. As a result, Qwen surpassed other models in this regard.

In summary, attribute discrepancies depended on the models' reasoning capabilities, parameter counts, and training methodologies.

*Results for Automated Deduplication Algorithm Tests*

We compared the traditional rule based deduplication with LLM based deduplication algorithm to figure out if LLM based deduplication has the potential to surpass rule based deduplication. A carefully controlled synthetic dataset was created containing news examples to do the comparisons. The synthetic dataset contained a total 462 examples including duplicates. The rule based deduplication dropped entries having duplicates in the fields: 'Accident Date', 'Killed', 'Injured' and 'Location'. The LLM based algorithm were tested using different LLMs. **Table-5** below shows the performance of deduplication algorithm with different LLMs.

**Table 5 Performance of Deduplication Algorithm Across Different Models**

| | **Rule Based** | **Gemini-2.5-flash** | **Gemma-3-27B-it-unsloth-bnb-4bit** | **Qwen-3-32B-AWQ** |
|---|---|---|---|---|
| Precision | 1 | 0.988 | 0.99 | 0.99 |
| Recall | 0.695 | 0.787 | 0.691 | 0.605 |
| F1 | 0.820 | 0.863 | 0.815 | 0.751 |

**Figure 11** below compares time taken to perform deduplication of the entire dataset with F1-score.

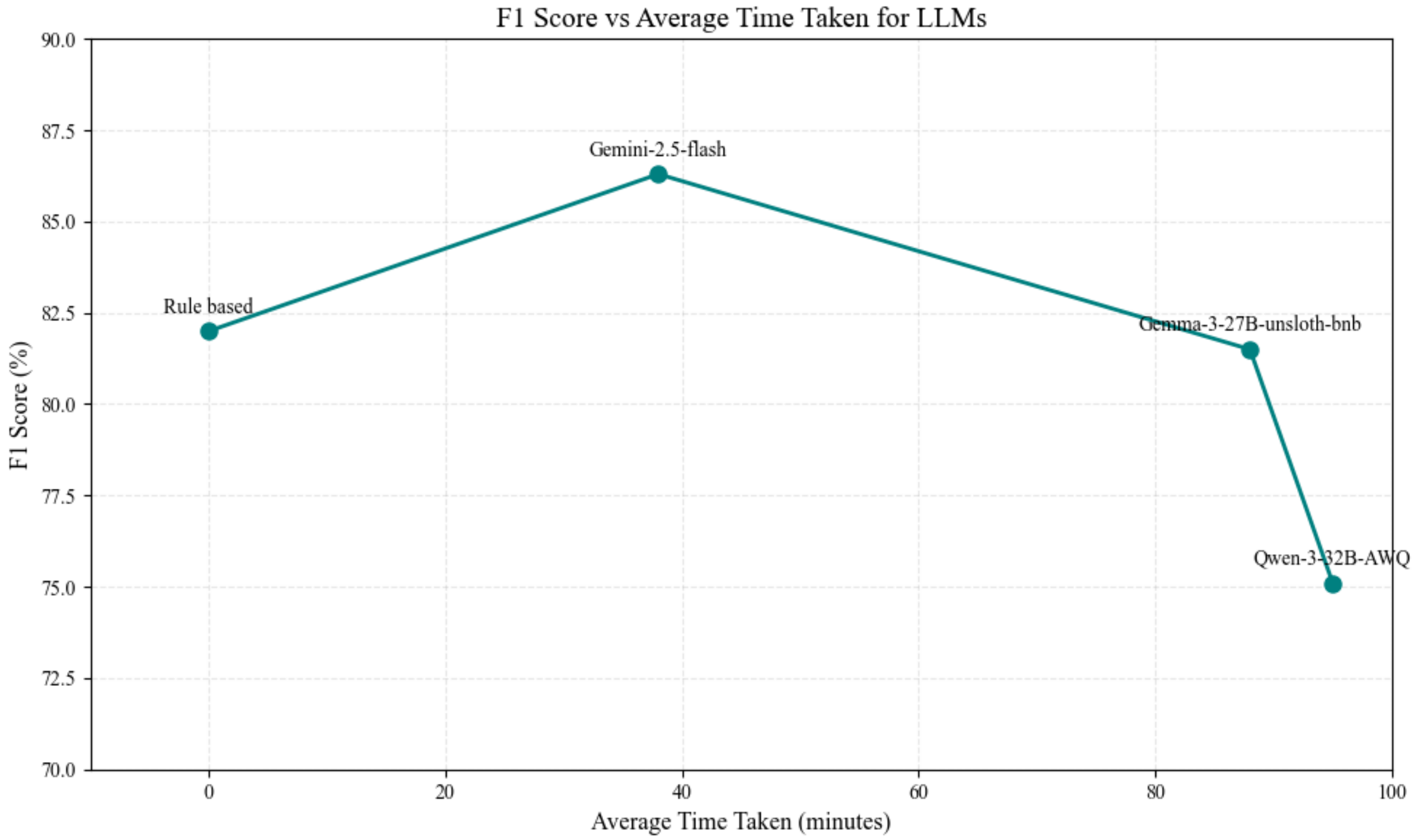

**Figure 11 F1-score(%) vs. Average Time Taken in Minutes for Deduplication**

It is can be seen that LLM powered deduplication can surpass traditional rule based deduplication in terms of F1-score accuracy. Particularly, LLM based algorithm can achieve higher Recall when used with higher parameter model. It is because, some news reports may contain inconsistent details of the same accident incident. Traditional rule based deduplication fails to capture these inconsistent duplicates. However, LLMs with competent reasoning capability can discern these duplicates. The downside, however, is that LLMs take quite a lot of time compared to rule based deduplication. However, if accuracy is our priority rather than time, LLM based deduplication can be a viable option in such cases.

*Time for Inference and Scalability*

The time for inference for processing each report is reported in the **Figure 12** below for different LLMs.

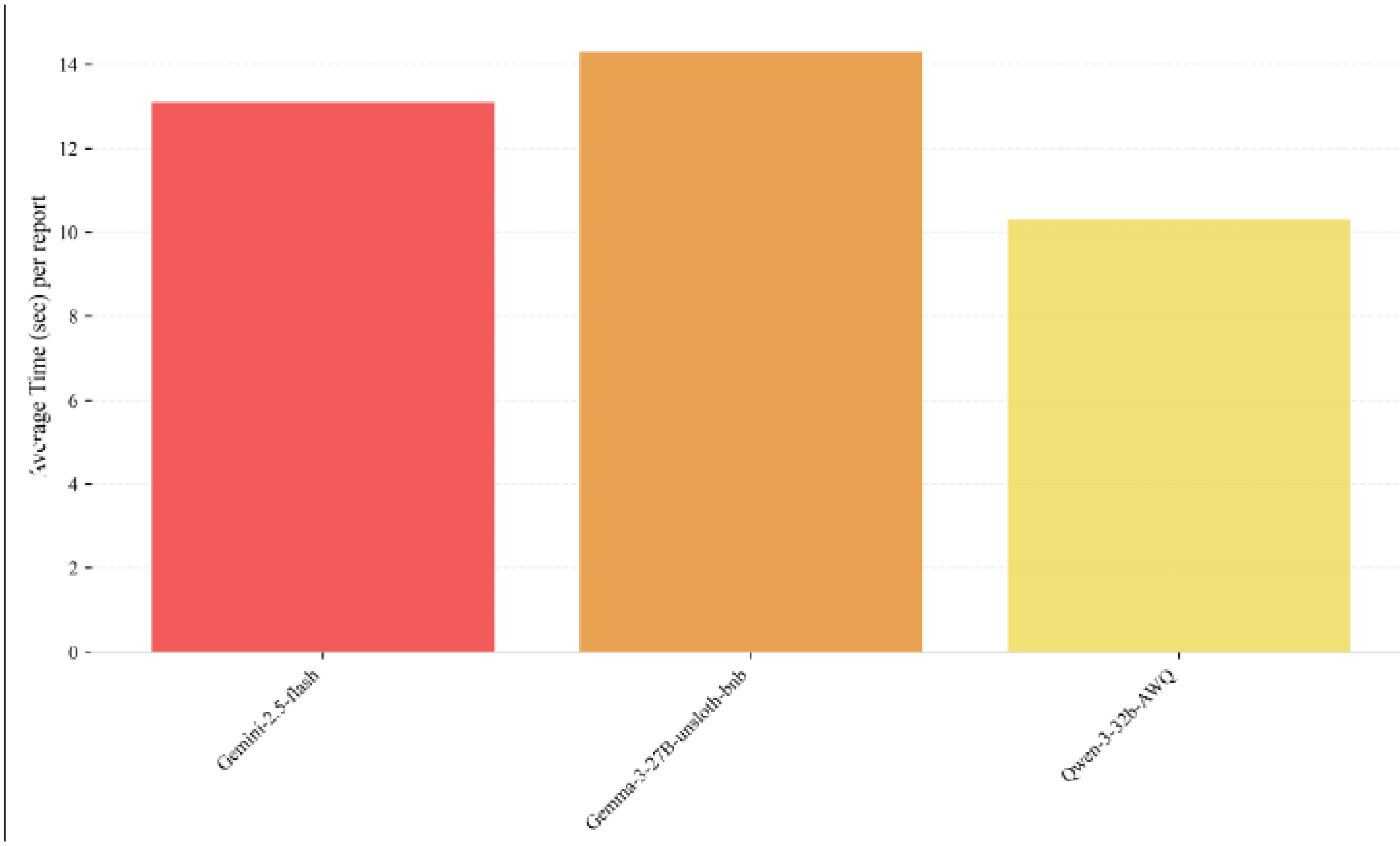

**Figure 12 Average Time Taken to Process Each Report for Different LLMs**

It can be seen that Qwen-3-32B-AWQ takes the lowest time to process each report as it has lower parameter count. Usually, LLMs with lower parameter count do inference faster. Gemini-2.5-flash comes in second and Gemma-3 takes the highest time.

In case of scalability, where thousands of news reports are to be processed, per 1000 reports would take 218 minutes, 238 minutes and 173 minutes respectively using a single instance each of these LLMs. If we use a single instance of the workflow, it will thus require a lot of time to process. However, if we perform horizontal scaling and distribute the processing task across multiple APIs or multiple GPUs, this time will significantly reduce. Using two instances will reduce the time by half and using three instances will reduce the time by one-third. Since, using GPUs for horizontal scaling is quite costly, we utilized API based scaling in the next section where we processed a large number of reports. We used 12 API instances simultaneously to distribute our processing tasks and thus our time for processing each 1000 reports decreased to 19 minutes only from 218 minutes.

*Accident Trend of Bangladesh Obtained Through Analysis of Data Collected Using This System*

The designed system was employed to collect accident data of Bangladesh from 2024-10-01 to 2025-01-20. So, accident data for a total of 111 days were collected from 14 different news sources. The system processed around 12,000 news reports in just 4 hours. The key takeaways from the analysis of the dataset is discussed in this section. We mainly used Gemini-2.5-flash API for this purpose as it offers better scalability than the open-source options.

Total number of accident occurrences during these 111 days is 705. **Figure 13, Figure 14** and **Figure 15** show the heatmaps generated after analyzing the data.

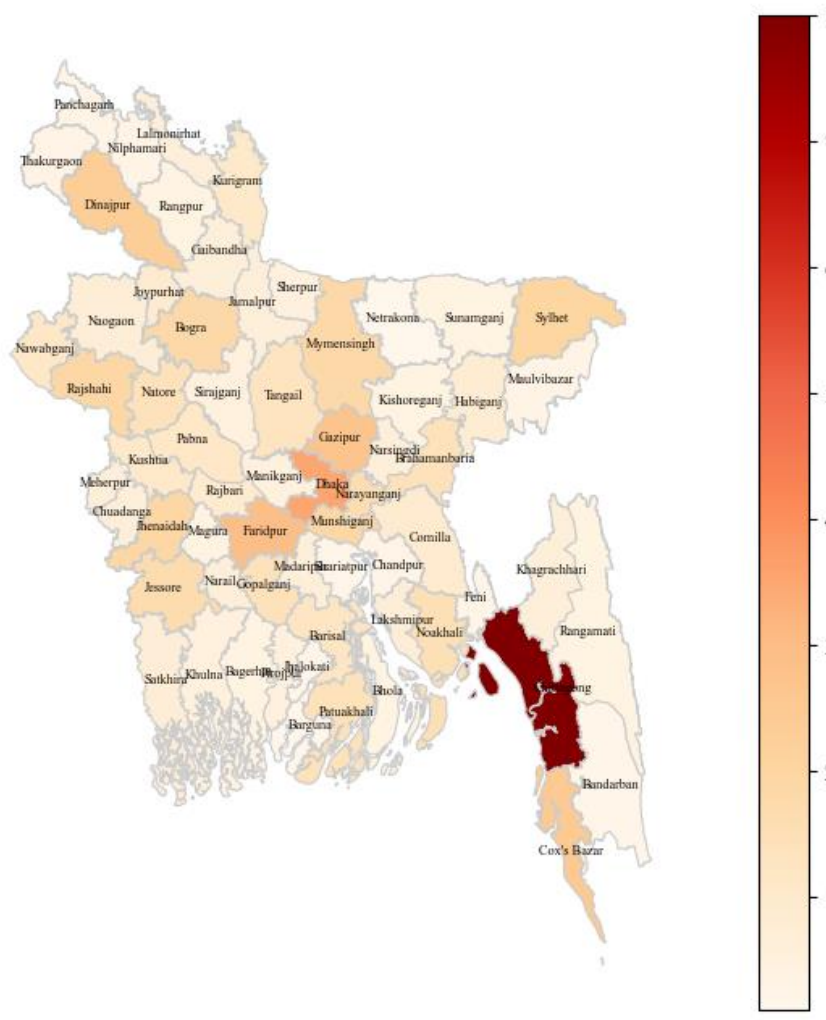

**Figure 13 Heatmap of Districtwise Road Traffic Crash Occurrences in Bangladesh**

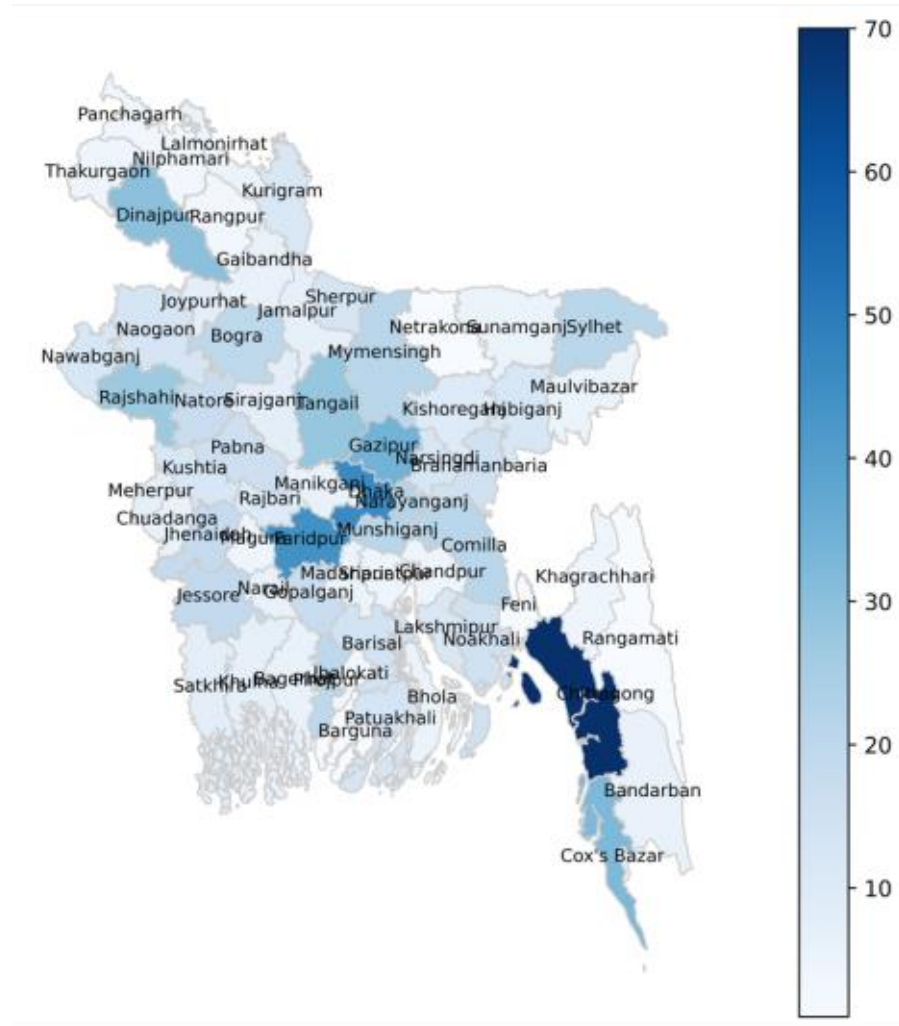

**Figure 14 Heatmap of District Wise Fatalities Due to Road Traffic Crashes in Bangladesh**

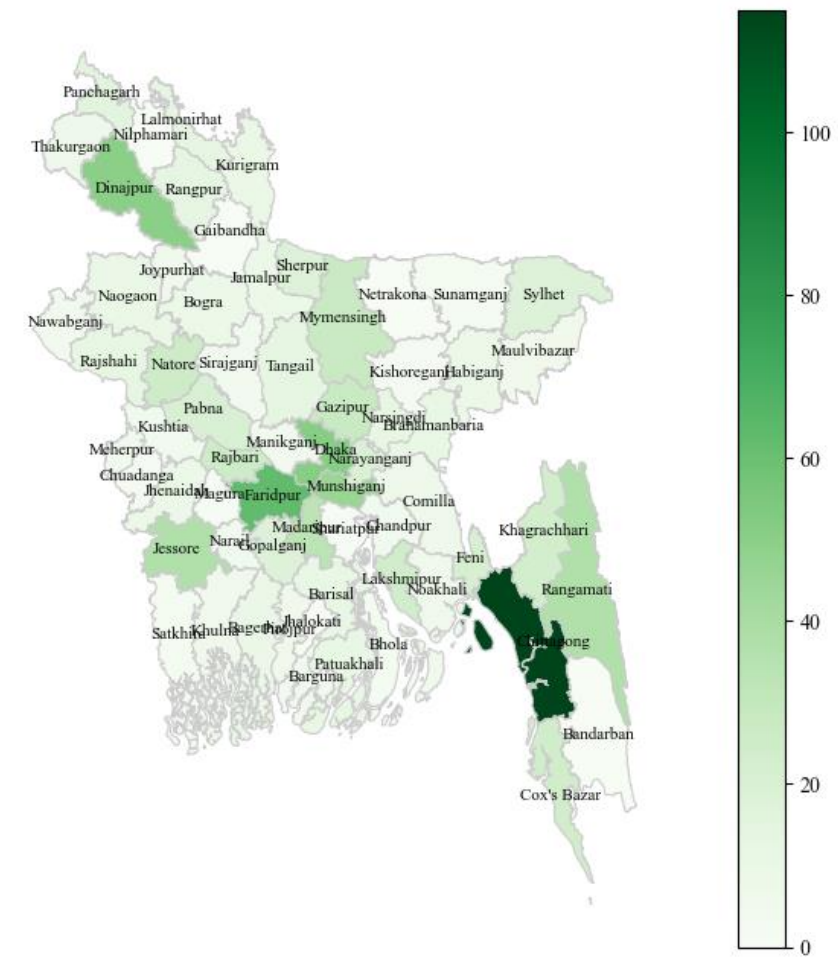

**Figure 15 Heatmap of District Wise Injuries due to Road Traffic Crashes in Bangladesh**

Several conclusions can be drawn from these figures. Chittagong emerges as the district with the highest frequency of road traffic accidents and fatalities, indicating significant road safety concerns, as it experienced 80 accidents resulting in 70 fatalities. Dhaka, despite having fewer accidents (35) compared to Gazipur (37), recorded a higher number of fatalities (46), suggesting that accidents in Dhaka might involve more severe outcomes or higher traffic densities leading to increased fatality rates.

*Miscellaneous*

We would like to draw attention to a report by World Bank. It says the following about crash data in Bangladesh.

*The current system of recording, analyzing, and reporting crashes in Bangladesh is cumbersome, error-prone, time-consuming, and is unsuitable for analysis and benchmarking. Crash data is characterized by irregular and unreliable year-on-year changes, indicating incompleteness and a lack of consistent procedures. Weak coordination between the ministries and inadequate internal organizational capacity result in crashes not being recorded and consequent severe underreporting of fatalities and injuries: a comparison of the recorded fatalities with the WHO estimated fatalities from 2010-2015 shows a discrepancy of more than 90 percent. Official crash fatality data from 2016 up to the present is still unavailable. Further, the Highway Police is invariably short of qualified and trained manpower, logistics, and equipment, including vehicles and other transport facilities, which affects its mobility. (3).*

So, in a country like Bangladesh, it extremely difficult even for researchers to get crash data from official sources such as- police. It is purely a socio-political problem rather than a technical one. Moreover, for the same reason, it is difficult to perform a ground truth validation as police records are simply not available to researchers and the available records are incomplete to the extent that it is not possible to reliably align the police records with newspaper records. So, despite having some under reporting bias of newspapers *(25)*, it is the best source to obtain accident data for transportation safety researchers in countries like Bangladesh. As a result, our designed system can pave the way for future transportation safety researches in similar contexts of Bangladesh.

## CONCLUSIONS

This research successfully demonstrates the transformative potential of Large Language Models (LLMs) in automating the generation of accident datasets, addressing a critical data gap in road safety management, particularly in resource-constrained nations like Bangladesh. The study introduces a novel, end-to-end pipeline that automates web scraping code generation, news classification, information extraction, and data deduplication, significantly minimizing manual effort and enhancing data accuracy and scalability. To our knowledge, this is the first study to automate both the generation of scraping scripts and the deduplication of accident reports using LLMs within the context of road traffic safety.
A key contribution of this work is its design for minimal local computational needs by leveraging generative LLMs, making it a viable solution for regions where advanced machine-learning infrastructure is scarce.
Future work should focus on incorporating more website types and creating hybrid models that combine generative LLMs with rule-based classifiers to improve accuracy and handle a wider variety of website architectures. Also, domain specific fine-tuning in addition to prompt engineering can be explored as well to increase accuracy in certain attributes. There is also significant potential in integrating real-time data sources like traffic cameras and police logs, which could be cross-verified by LLMs to create a near real-time accident alert system.
In conclusion, by transforming unstructured news data into a valuable, structured format, this research provides a robust foundation for data-driven policymaking. The proposed system offers a scalable and adaptable approach to enhance road safety measures in Bangladesh and can be extended to other data-scarce environments and different types of incident tracking.

**ACKNOWLEDGEMENTS**
ChatGPT (GPT-4o) was used to organize the paper’s language and grammar only. It was ensured that AI contents do not contain any plagiarism.

**AUTHOR CONTRIBUTIONS**
MD Thamed Bin Zaman Chowdhury: Conceptualization, Data curation, Formal analysis, Investigation, Methodology, Resources, Software, Validation, Writing – original draft
Moazzem Hossain: Validation, Supervision, Methodology, Conceptualization.

**DECLARATION OF CONFLICTING INTERESTS**
The authors declare that they have no known competing financial interests or personal relationships that could have appeared to influence the work reported in this paper.

**DATA AVAILABILITY**
Data will be made available on request.

**REFERENCES**
1. Association for Computing Machinery (ACM). (2018). ACM Code of Ethics and Professional Conduct. https://www.acm.org/code-of-ethics

2. Cheng, P., Xiao, W., Ning, P., Li, L., Rao, Z., Yang, L., Schwebel, D. C., Yang, Y., Huang, Y., & Hu, G. (2022). ARTCDP: An automated data platform for monitoring emerging patterns concerning road traffic crashes in China. Accident Analysis & Prevention, 174, 106727. https://doi.org/10.1016/j.aap.2022.106727

3. Daniels, S., Brijs, T., & Keunen, D. (2010). Official reporting and newspaper coverage of road crashes: A case study. Safety Science, 48(10), 1469–1476. https://doi.org/10.1016/j.ssci.2010.07.007

4. Dong, Q., Li, L., Dai, D., Zheng, C., Ma, J., Li, R., Xia, H., Xu, J., Wu, Z., Chang, B., Sun, X., Li, L., & Sui, Z. (2024). A survey on in-context learning. In Y. Al-Onaizan, M. Bansal, & Y.-N. Chen (Eds.),

Proceedings of the 2024 Conference on Empirical Methods in Natural Language Processing (pp. 1107–1128). Association for Computational Linguistics. https://doi.org/10.18653/v1/2024.emnlp-main.64

5. Gemma Team. (2025). Gemma 3. Kaggle. https://goo.gle/Gemma3Report

6. Google AI. (n.d.). Gemini models. Retrieved March 15, 2025, from https://ai.google.dev/gemini-api/docs/models/gemini

7. Google Cloud. (n.d.). Vertex AI documentation. Retrieved March 15, 2025, from https://cloud.google.com/vertex-ai/docs

8. Hugging Face. (2020). Transformers: State-of-the-art natural language processing for Pytorch, TensorFlow, and JAX. https://huggingface.co/docs/transformers

9. Jaradat, S., Nayak, R., Paz, A., Ashqar, H. I., & Elhenawy, M. (2024). Multitask learning for crash analysis: A fine-tuned LLM framework using Twitter data. Smart Cities, 7(5), 2422–2465.

10. Metreau, E., Young, K. E., & Eapen, S. G. (2024, July 1). World Bank country classifications by income level for 2024–2025. World Bank Blogs. https://blogs.worldbank.org/en/opendata/world-bank-country-classifications-by-income-level-for-2024-2025

11. NVIDIA Corporation. (2023). CUDA Toolkit Documentation. https://developer.nvidia.com/cuda-toolkit

12. Oliaee, A. H., Das, S., Liu, J., & Rahman, M. A. (2023). Using Bidirectional Encoder Representations from Transformers (BERT) to classify traffic crash severity types. Natural Language Processing Journal, 3, 100007.

13. Paszke, A., Gross, S., Massa, F., Lerer, A., Bradbury, J., Chanan, G., Killeen, T., Lin, Z., Gimelshein, N., Antiga, L., Desmaison, A., Kopf, A., Yang, E., DeVito, Z., Raison, M., Tejani, A., Chilamkurthy, S., Steiner, B., Fang, L., Bai, J., & Chintala, S. (2019). PyTorch: An imperative style, high-performance deep learning library. Advances in Neural Information Processing Systems, 32, 8024–8035. https://papers.nips.cc/paper_files/paper/2019/hash/bdbca288fee7f92f2bfa9f7012727740-Abstract.html

14. Qwen Team. (2025). Qwen3 Technical Report. arXiv. https://arxiv.org/abs/2505.09388

15. Reitz, K. (2023). Requests: HTTP for humans™. Read the Docs. Retrieved March 15, 2025, from https://docs.python-requests.org/en/latest/index.html

16. Richardson, L. (2004). Beautiful Soup. Retrieved March 15, 2025, from https://launchpad.net/beautifulsoup

17. Selenium Contributors. (n.d.). WebDriver documentation. Selenium. Retrieved March 15, 2025, from https://www.selenium.dev/documentation/webdriver/

18. Surendra, A., Sood, K., Albuquerque, C., Akbar, S., Joshi, G., & Sakshi, P. (2024). Revolutionizing Road Accident Analysis: A Deep Learning Approach to Entity Recognition and Extraction. 2024 IEEE 15th Annual Ubiquitous Computing, Electronics & Mobile Communication Conference (UEMCON), 308–314. https://doi.org/10.1109/UEMCON62879.2024.10754665

19. Tillet, P., Zheng, H., & Gholami, A. (2021). Triton: An intermediate language and compiler for tiled neural network computations. https://github.com/openai/triton

20. Wang, X., Chen, Y., Yuan, L., Zhang, Y., Li, Y., Peng, H., & Heng, J. (2024). CodeAct: Your LLM agent acts better when generating code. In Proceedings of the 41st International Conference on Machine Learning (ICML).

21. World Bank. (2022). Bangladesh Road Safety Project. Retrieved March 15, 2025, from https://documents1.worldbank.org/curated/en/315861648600487321/pdf/Bangladesh-Road-Safety-Project.pdf

22. World Health Organization. (2018). Global status report on road safety 2018. World Health Organization. Retrieved March 15, 2025, from https://iris.who.int/handle/10665/276462

23. Yang, X., Bekoulis, G., & Deligiannis, N. (2023). Traffic event detection as a slot filling problem. Engineering Applications of Artificial Intelligence, 123, 106202. https://doi.org/10.1016/j.engappai.2023.106202

24. Yuan, S., & Wang, Q. (2022). Imbalanced Traffic Accident Text Classification Based on Bert-RCNN. Journal of Physics: Conference Series, 2170(1), 012003. https://doi.org/10.1088/1742-6596/2170/1/012003

25. Zhen, H., Shi, Y., Huang, Y., Yang, J. J., & Liu, N. (2024). Leveraging large language models with chain-of-thought and prompt engineering for traffic crash severity analysis and inference. Computers, 13(9), 232.